\documentclass{article}

\usepackage{hyperref}       % hyperlinks
\usepackage[square,numbers]{natbib}

\usepackage[final]{nips_2016}
\usepackage{algpseudocode} 
\usepackage{algorithm}
\usepackage[utf8]{inputenc} % allow utf-8 input
\usepackage[T1]{fontenc}    % use 8-bit T1 fonts
\usepackage{url}            % simple URL typesetting
\usepackage{booktabs}       % professional-quality tables
\usepackage{amsfonts}       % blackboard math symbols
\usepackage{nicefrac}       % compact symbols for 1/2, etc.
\usepackage{microtype}      % microtypography
\usepackage{mathtools}
\usepackage{amsmath}
\usepackage{caption}
\usepackage{subcaption}
\newcommand{\head}[1]{\noindent \textbf{#1}}

\title{Learning Runtime Parameters in Computer Systems with Delayed Experience Injection}

  \author{
  Michael Schaarschmidt \\
  University of Cambridge\\ 
  \texttt{michael.schaarschmidt@cl.cam.ac.uk} \\
   \And
Felix Gessert \\
  University of Hamburg \\
  \texttt{gessert@informatik.uni-hamburg.de} \\
 \And
 Valentin Dalibard \\
  University of Cambridge\\ 
  \texttt{valentin.dalibard@cl.cam.ac.uk} \\
 \And
Eiko Yoneki \\
  University of Cambridge\\ 
  \texttt{eiko.yoneki@cl.cam.ac.uk} \\
}

\begin{document}

\maketitle

\begin{abstract}
Learning effective configurations in computer systems without hand-crafting models for every parameter is a long-standing problem. This paper investigates the use of deep reinforcement learning for runtime parameters of cloud databases under latency constraints. Cloud services serve up to thousands of concurrent requests per second and can adjust critical parameters by leveraging performance metrics. In this work, we use continuous deep reinforcement learning to learn optimal cache expirations for HTTP caching in content delivery networks. To this end, we introduce a technique for asynchronous experience management called delayed experience injection, which facilitates delayed reward and next-state computation in concurrent environments where measurements are not immediately available. Evaluation results show that our approach based on normalized advantage functions and asynchronous CPU-only training outperforms a statistical estimator. 
\end{abstract}

% should we motivate the configuration problem here?
\section{Introduction}
In recent years, reinforcement learning (RL) algorithms have been successfully combined with deep neural networks as function approximators \citep{MnihKavukcuogluSilverEtAl2013, MnihDQN2015, 2015arXiv150906461V}. Neural networks can capture structure in the environment from high-dimensional raw inputs and efficiently generalize over large state spaces. Deep reinforcement learning (DRL) techniques hence provide a powerful end-to-end task learning model from sensory inputs without prior knowledge of environment dynamics. However, training value functions for complex tasks requires significant training times and substantial computational resources.

There is another set of control problems in the domain of computer systems which are characterized by smaller problem dimensions and strong latency constraints. The problem this paper addresses is the utilization of DRL to provide real-time controllers for such problems. The key idea of this paper is that for comparatively small state dimensions (< 100) and tasks with weaker structure, no extended offline training is necessary to implement effective controllers.

As an example application, we consider cloud database services (database-as-a-service; DBaaS), which manage data and automate the operations of distributed database infrastructures. They typically employ a convention-over-configuration paradigm and do not adjust request-level parameters unless specified by developers.
Nonetheless, many configuration parameters have significant performance impact for clients. In order to adjust them at runtime, one must address several challenges. 

First, the impact of individual actions taken in the system is difficult to measure due to serving many concurrent requests. Further, measuring system performance might only be possible after some time has passed. Second, client-server architectures might prevent infrastructure providers from directly observing client performance. 
This work addresses the challenges of concurrent delayed credit assignment by introducing a mechanism for concurrent asynchronous experience management called delayed experience injection. Specifically, we modify normalized advantage functions \citep{GuLillicrapSutskeverEtAl2016}, a recently introduced method for continuous deep reinforcement learning, to learn optimal cache expiration durations for dynamically changing query results. Results show that our controller outperforms a statistical estimator based on arrival processes.

\section{Background and related work}
\subsection{Preliminaries}
The parameter learning problem conforms to the setting of an infinite-horizon discounted Markov decision process where an agent interacts with an environment described by states $s \in \mathcal{S}$ and aims to learn a policy $\pi$ that governs which action $a \in \mathcal{A}$ to take in each state \citep{SuttonBarto1998}. At each discrete time step $t$, the agent takes an action $a_t$ according to its current policy $\pi(a|s)$, transitions into a new state $s_{t+1}$ according to the (often stochastic) environment dynamics, and observes a reward $r_t$. The goal of the agent is to maximize cumulative expected rewards $R = \mathbb{E}[\sum_t \gamma^{t}r_t]$, where future rewards are discounted by $\gamma$. This is often achieved by learning a Q-function $Q^{\pi}(s_t,a_t)$ which represents the expected return when starting from state $s_t$, taking action $a_t$ with the highest Q-value and following $\pi$  thereafter \citep{WatkinsDayan1992}. Mnih et al. have demonstrated how deep neural networks can be used as value functions for a variety of complex tasks by utilising a replay memory of stored experiences, and using a second value function to stabilize learning (fixed Q-target)~\citep{MnihDQN2015}.

In this work, we utilize normalized advantage functions (NAFs), which have recently been suggested as an effective method for continuous DRL \citep{GuLillicrapSutskeverEtAl2016}. The key problem in continuous RL is to efficiently select the action maximising the Q-function, i.e. $arg~max_aQ(s,a)$ while avoiding to perform a costly numerical optimization at each step. Unlike other approaches in continuous DRL (e.g. deep deterministic policy gradients \citep{LillicrapHuntPritzelEtAl2015}), NAFs avoid the use of a second actor or policy network that needs to be trained separately. A single neural network $Q(s,a|\theta^Q)$ is used to output both a value function $V(s|\theta^V)$:
\begin{align}
V^{\pi}(s_t|\theta^V) = \mathbb{E}_{r_{i\geq t},s_{i>t}\sim E,a_{i\geq t}\sim\pi}[R_t|s_t,a_t]
\end{align}
and an an advantage term $A^\pi(s_t,a_t)$:
\begin{align}
A^{\pi}(s_t,a_t|\theta^A) = Q^\pi(s_t,a_t|\theta^Q) - V^{\pi}(s_t|\theta^V)
\end{align}
Decomposing $Q$ into a state-value term $V$ and an advantage term $A$ is a technique for variance reduction often used in policy gradient methods \citep{1993b, 1994hbk}. Gu et al. suggest using a quadratic advantage term:
\begin{align}
A(s,a|\theta^A) = -\frac{1}{2}(a - \mu(s|\theta^{\mu}))P(s|\theta^P)(a - \mu(s|\theta^{\mu})),
\end{align}
where $P(s|\theta^P)$ is a positive-definite square matrix parametrized by a lower-triangular matrix $L(s|\theta^P)$, which is given by a linear output of the network ($P(s|\theta^P)=L(s|\theta^P)L(s|\theta^P)^T$), with the diagonal entries exponentiated. Hence, the maximizing action is always given by  $\mu(s|\theta^\mu)$. Updates are computed by minimizing the mini batch loss $L = \frac{1}{N}\sum_i (\gamma_i - Q(s_i,a_i|\theta^Q))^2$ and using a replay memory, as well as a target network $Q'$ (as described by Mnih et al.) to compute $y_i = r_i + \gamma V'(s_{i+1}|\theta^{Q'}) $. NAFs are especially appealing in our context because using a single network simplifies asynchronous update semantics.

\subsection{Related work}
Our work is conceptually most similar to Tesauro et al.'s work on resource allocation in data centres~\citep{TesauroDasWalshEtAl2005,TesauroJongDasEtAl2006, Tesauroothers2005}. They utilized a perceptron with a single hidden layer to make server allocation decisions for different applications. Their method aims to maximize the expected sum of service level agreement payments while minimizing penalties for unmet service-level objectives. Their state comprised the mean arrival rate of HTTP requests and the number of currently allocated servers. The same approach has also been successfully applied to power management of web servers \citep{TesauroDasChanEtAl2007}. Notably, their solution relies on a hybrid approach where initial values are improved by a parametric model. Our work similarly relies on arrival rates of certain events but shifts learning from global state and server-level decisions to per-request state and request-level decisions. RL has also been employed for auto-configuration of Xen virtual machines \citep{RaoBuXuEtAl2011, XuRaoBu2012}. 

For web caching, Candan et al. initially explored the notion of invalidation-based caching for web content \citep{candan_enabling_2001}, as opposed to treating web caches as static content stores or media distribution servers \citep{Freedman2010, HuangBirmanRenesseEtAl2013}. 

Prior approaches on thread-parallel or distributed DRL such as A3C \citep{MnihBadiaMirzaEtAl2016} or Gorila \citep{NairSrinivasanBlackwellEtAl2015} accelerate training by having learners operate on separate copies of single-threaded environments (e.g. Atari simulator). Gu et al. have also recently applied distributed asynchronous NAFs to shared learning of 3D robot manipulation tasks \citep{2016arXiv161000633G}. In their work, distributed robot controllers asynchronously share their (sequentially) collected experiences with a central server. In contrast, our work considers thread-asynchronous training in a single node environment with a high degree of concurrency and delayed asynchronous reward assignment.

\section{Problem overview}
\subsection{Estimating cache expirations}
We consider the problem of learning parameters for cloud database services on a per-request level granularity. For each request, the database server can set response parameters affecting client performance. Multiple clients (e.g. mobile devices) can query and update the same entries in a single database. In this paper, we address the problem of estimating cache expiration times (time-to-live; TTL) for dynamically changing query results, which we now introduce. 

A query $q$ issued by a client is executed by a database and yields a set of result records of varying cardinality $n$, identified by their unique keys $k_1,..,k_n$. Query results can be cached for a specified time interval $t=TTL$ at server-controlled caches such as content delivery networks (CDNs) or reverse-proxy caches. If a key $k$ is updated, all cached queries containing $k$ become invalid and an invalidation request is sent to all caches. There are multiple reasons why estimating accurate TTLs for query results is critical. 

First, the server has to store all cached queries and their expiration times to determine which queries need to be invalidated. Using indiscriminately large TTLs for dynamically changing database content would thus both strain cache capacities as well as create too much overhead to determine invalidations, as every update needs to be compared against all cached queries. Further, every invalidation creates the potential for stale reads, as clients can retrieve stale cached results while the invalidation is propagated to all cache edges \citep{GessertSchaarschmidtWingerathEtAl2015}. In contrast, small TTLs increase client latencies significantly if the database server is physically remote since web performance is primarily governed by round-trip latency \citep{grigorik_high_2013}. In the following subsection, we will introduce a Monte Carlo framework designed to analyze web request flows.

\subsection{Simulation environment}
We have implemented a Monte Carlo simulation inspired by the Yahoo! cloud serving benchmark (YCSB) \citep{CooperSilbersteinTamEtAl2010, Patil:2011:YBP:2038916.2038925}. YCSB is a benchmark suite for cloud databases and defines a set of typical web workloads (e.g. read-dominant, scan-intensive, write-dominant). Custom workloads can be specified with properties such as request distribution, record count, operation count, and read/write/insert/scan/delete mixture. After running a workload, YCSB provides throughput and latency histograms. Our implementation provides the same workloads but instead of just providing a client interface and workloads, we stack together multiple layers (clients, caches, databases) and replicate web connection semantics. That is, YCSB operates on a synchronous thread-per-request model while web browsers typically use 6 HTTP connections and fetch multiple resources asynchronously.  

\begin{figure}[t!]
    \centering
     \includegraphics[scale=0.5]{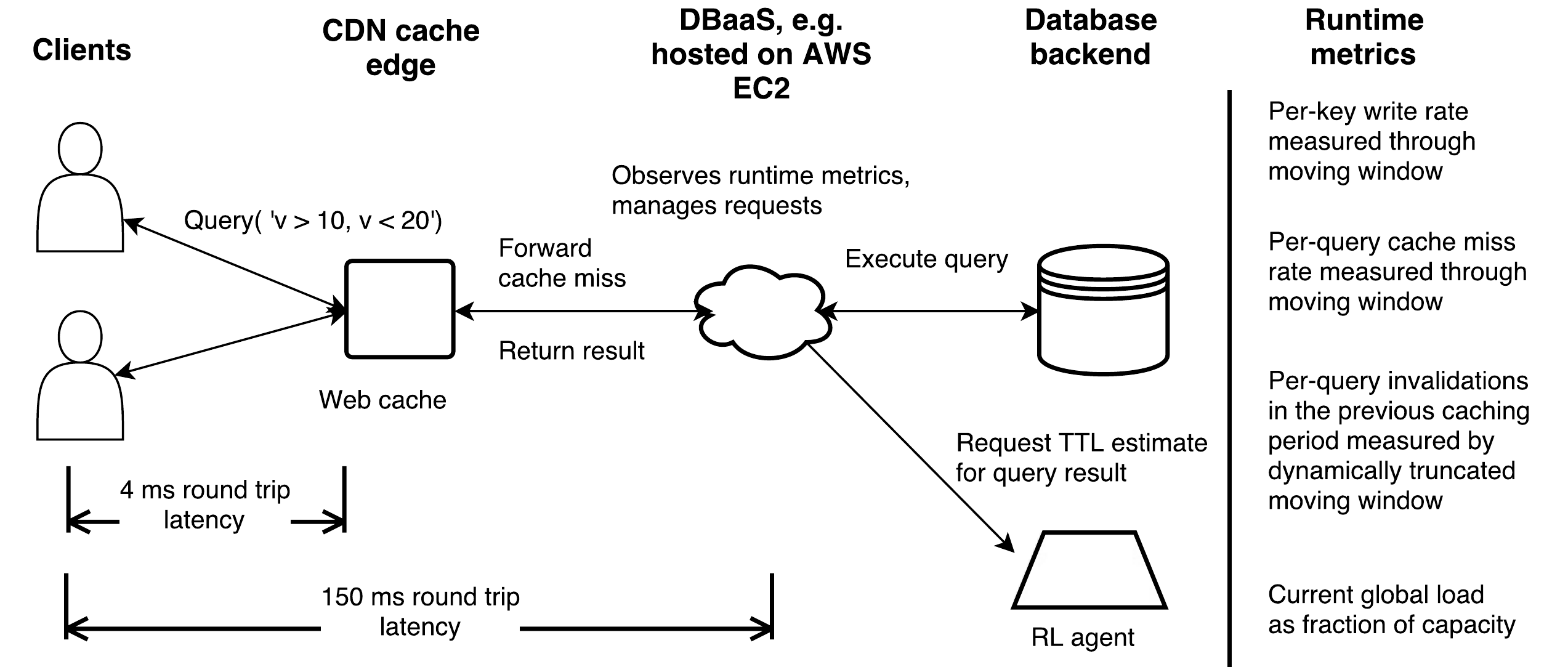}
  \caption{Overview of our Monte Carlo simulator. Clients issue requests which first go against local cache edges. On a cache miss, requests are forwarded to a backend in another geographical region, inducing much higher round-trip latency. The DBaaS middle ware collects metrics on writes, cache misses and invalidations.}
  \label{fig:simulation-query}
  \vspace{-5mm}
\end{figure}

Figure \ref{fig:simulation-query} gives a schematic overview of the request flow for queries. Clients sample query or update requests from the workload mixture. In the simulation, each entry has a single field with a numerical value. Operations read or modify a single key $k$ drawn from an access distribution. For easier result size control, queries are defined as range queries that request all objects for which the corresponding database entry satisfies the range predicate on the numerical field. 

We assume the setting of a geographically remote database server hosted in the California Amazon EC2 region with a client located in Europe. Clients can drastically reduce request latencies if dynamically changing query results are present in a near cache such as content delivery network (CDN) edge. Multiple clients may query and write the same data, e.g. by commenting on a social media post or refreshing their news feed. If a client executes an update operation on a key against the DBaaS, it determines which queries need to be invalidated by re-evaluating a maintained index of cached queries against the update (e.g. through stream processing). Entries are removed from the index once the respective TTL expires. In the simulation, we pre-construct an index of queries and initial result keys and can thus cheaply determine invalidated queries by incrementally updating this index at runtime. The DBaaS then sends out asynchronous invalidation requests to the CDN. We regard a read operation as a special case of query with result size one. If another client requests a cached entry before an invalidation has been completed, a stale read occurs~\citep{GessertSchaarschmidtWingerathEtAl2015}. For TTL estimation, the server can utilize update rates on records as well as cache miss rates and invalidations on queries. In the following section, we will discuss different TTL estimation strategies and explain how our approach leverages these metrics.
% mention cache capacity as motivation

\section{Estimating TTLs}
\subsection{True TTL}
We begin by considering a hypothetical optimal strategy. Ideally, TTLs are estimated to expire right before an update invalidates the respective cached result. We define the \textit{true TTL} as the interval between serving the query and the query result being invalidated by a write $w$. In our simulation, we can hence capture the optimal action for every step after the respective query has been invalidated.
% For every incoming request, we initially capture a tuple of $(state, action, reward, predicted TTL)$. When a query is invalidated (potentially after many further cache misses), we update all existing tuples for a query for which the true TTL has not been set based on arrival and invalidation timestamps. These traces hence allow us to evaluate solutions based on their (root-mean-squared) per-step error. 
Since this would not capture true TTLs for queries which expire from the cache without invalidation, we further measure the "theoretical" true TTL for queries which are currently not cached by evaluating which queries would have been invalidated if they had not expired. 

\subsection{Baseline solution}
We first introduce a baseline solution relying on the assumption of a Poisson process of incoming updates. For a Poisson process, the inter-arrival times of events have an exponential cumulative distribution function (CDF), i.e. each of the identically and independently distributed random variables $X_i$ has the cumulative density 
$F(x;\lambda) = 1 -  e^{(-\lambda x)}~for~x\geq0$
and mean $1/\lambda$. For now, we make the impractical assumption that for each database record, there is an estimate of the rate of incoming writes $\lambda_w$ over some time window.  

The result set of a query of cardinality $n$ can then be regarded as a set of independent exponentially distributed random variables $X_i,\ldots,X_n$ with different write-rates $\lambda_{w1}, \ldots, \lambda_{wn}$. Estimating the TTL for the next update to any element of the result set requires a distribution that models the minimum time to the next write, i.e. $X_{min} = min\{X_1,\ldots, X_n\}$, which is again exponentially distributed with $\lambda_{min} = \lambda_{w1} + \ldots + \lambda_{wn}$. We can hence obtain an estimate of the TTL by using the expected value until the next write on any record present in the result set, which would invalidate the cached result: $ TTL_{poisson} =\mathbb{E}[X_{min}] = 1/\lambda_{min} $. As we will show in the evaluation, the key problem of this approach is providing it with default write rates or default TTLs if no write-rate information is available. 

\subsection{TTL estimation with NAFs}
\head{Motivation.} TTL estimation is an appealing problem for reinforcement learning solutions as they provide a natural way to deal with time-dependent and noisy feedback loops in control problems. We proceed to model the TTL estimation problem using NAFs. First, the previous solution does not distinguish between queries that are read often and queries that are requested very rarely, i.e. it does not incorporate cache miss rates. Second, the baseline solution cannot deal well with sparse information in the state: For most objects, write rate information might not be available. Given no (or often partial) information an estimator needs to fall back to default values. 

\head{State.} We use individual record metrics to learn TTLs for query result sets. This is preferable to using an encoding of a query itself as the state, since many equivalent query strings lead to the same result. Using record-level metrics allows for an easier generalization when the result sets of seen and unseen queries overlap. Since update and query operations are generally independent, we also utilize query cache miss rates as part of the state to measure TTL impact by inputting the difference between current and last miss rate. 

Query results can significantly vary in size but the contribution of records which are rarely updated to the TTL should be negligible. We hence set the number of inputs to the mean expected result size $n$ and input a sorted vector of the top $n$ available write rates to the network -- other components in the case of $card(result) > n$ are discarded. 

\head{Reward.} 
The reward needs to encode as much information as possible from what the DBaaS can observe. From a service provider's perspective, rewards should allow to trade off invalidations against cache misses. The server cannot observe direct reward measures such as cache hit rates for clients or CDN cache utilisation. We note that we expect most queries not to be invalidated frequently (or at all) due to the power-law nature of web workloads \citep{BreslauCaoFanEtAl1999}. Hence, using the expected invalidation rate as a TTL estimation strategy is unlikely to be successful as there will be no information for most queries.

To punish invalidated queries, the agent needs to know which actions cause a query to be invalidated by a later write. This means the server can only sensibly measure a reward after some delay $t_d$. The same problem exists for state measurements relying on cache miss rates. If there is an invalidation at time $t_{inv}$ before the expiration timestamp of a cached query $t_{exp}$, the reward can be computed as the difference between invalidation time and expiration time (in seconds), i.e. $r_t=t_{inv} - t_{exp}$. If there has been no invalidation, less informative metrics have to be used. 

No invalidation before expiration means that the TTL  could have been higher unless capacity constraints prohibit longer caching times. In this case, we hence use a static reward $r$ and scale it by the current load $c_t$ (current cached queries divided by capacity) to encourage longer TTLs when fewer queries are cached and shorter TTLs when load is close to capacity (by using $-c_t$ if larger than some threshold), i.e. $r_t = r \cdot (1 + c_t)$. 
The intuition behind this approach is that only using invalidation timestamps would not allow to give a reward for $80-90\%$ of queries (due to low invalidation rates, as shown in the evaluation), and would not give opportunity to globally up- or down-regulate estimates according to system-wide load. In the following section, we explain how we practically perform delayed reward and next-state measurements.
%Finally, we note that our simulation would also allow to practically explore Gu et al.'s interpretation of Q-learning as variational inference by using the hypothetical true TTL for expired queries which would have been invalidated upon later writes as the loss and learning fully on policy, which is beyond the scope of this paper.

\subsection{Delayed experience injection}
In standard RL semantics, the agent sequentially moves through a Markov decision process by taking steps and recording transitions of state, action, reward and next-state. When using a replay memory, learning is decoupled from current state and actions by sampling transitions from the memory to perform mini-batch gradient descent. Consequently, if the desired runtime measurements for rewards and next-states are not available immediately and the agent has to deal with many concurrent requests, the application needs to keep track of "incomplete" transitions and decide when to complete them. 

\begin{algorithm}
\begin{algorithmic}
\State Initialize empty replay memory $\mathcal{R} \gets \emptyset$
\State Initialize Q-network $Q(s,\mu|\theta^Q)$ with random weights
\State Initialize target network Q' with weight $\theta^{Q'} \gets \theta^Q$
\State Initialize random process $\mathcal{N}$ for initial exploration
\For{ $t=1, T$}
\State Select action $a_t = \mu(s_t|\theta^{\mu}) + \mathcal{N}_t$
\State Create incomplete transition $(s_t,a_t)$, \State Enqueue $(s_t,a_t)$ in expiration queue with $exp_t= now() + t_d$
\State At $t=exp_t$, asynchronously execute queue consumer: 
\Statex\hskip\algorithmicindent\hskip\algorithmicindent   Compute $r_t$ and $s_{t+1}$ 
\Statex \hskip\algorithmicindent\hskip\algorithmicindent Insert complete transition $(s_t,a_t,r_t,s_{t+1})$ into $\mathcal{R}$.
\State Submit asynchronous loss computation:
\Statex \hskip\algorithmicindent\hskip\algorithmicindent Compute $y_i = r_i + \gamma V'(s_{i+1}|\theta^{Q'}) $
\Statex \hskip\algorithmicindent\hskip\algorithmicindent Minimize $L = \frac{1}{N}\sum_i (\gamma_i - Q(s_i,a_i|\theta^Q))^2$
\Statex\hskip\algorithmicindent\hskip\algorithmicindent Periodically update $\theta^{Q'} \gets \theta^Q$
\EndFor
\end{algorithmic}
\caption{Asynchronous NAF with delayed experience injection.}
\label{async-naf}
\end{algorithm}

Algorithm \ref{async-naf} shows the control flow in our model. The DBaaS server computes the state from write rate and cache miss metrics for an incoming query and creates an incomplete transition $(s_t,a_t)$. This is then enqueued into an \textit{expiration queue} data structure which triggers an asynchronous consumer after the specified delay $t_d$, which we set to $a_t$ in our experiments (i.e. the TTL). The consumer computes the reward and the next state as described above by requesting the last invalidation timestamp and cache miss rate. It then inserts the completed transition $(s_t,a_t,r_t,s_{t+1})$ into the replay memory $\mathcal{R}$, a mechanism we call \textit{delayed experience injection} (DEI). DEI decouples not only current state from learning (as a replay memory does) but also decouples future state and reward computation for specific queries from the sequence of incoming states.  Hence, NAF-DEI also solves a different problem than the recently introduced distributed asynchronous NAF \citep{2016arXiv161000633G}, where multiple controllers sequentially collect experiences without delay in the experience computation itself. Further, the difference between DEI and the standard delayed reward assignment problem \citep{Watkins:1989} is that DEI deals with concurrent delayed credit assignment. 

Updates are performed similar to standard NAF except that the update step is also computed asynchronously by another thread. This is necessary because blocking incoming decision queries on the update step would result in latency spikes. 

\section{Evaluation}
\subsection{Setup}
The goal of the evaluation is to demonstrate the principal feasibility of using deep reinforcement learning for request-level parameter learning. We begin by describing the experimental setup. We have implemented our simulator in Java 8 (for YCSB compatibility) and utilized deeplearning4j \citep{dl4j} (0.5.0) for the NAF implementation. We set 10 clients with each 6 concurrent connections to execute a combined target throughput of 1,000 (asynchronous) operations per second. They accessed 10,000 documents with 1,000 distinct queries under varying workloads. Updates and queries were drawn from a Zipfian distribution (Zipf constant $0.6$). Each workload was run for 30 minutes on a commodity 4 core desktop machine and results were averaged over five runs. Query result sizes were set to be between $[1,20]$ documents by sampling scan ranges from $\mathcal{N}(10,5)$ (resp. $\mathcal{N}(5, 2)$ in smaller experiments). Simulated round-trip latencies reflected a client in Europe, a CDN edge in Europe (4 ms round trip latency), and a server in the EC2 California region (150 ms round trip latency.)

The NAF agent used 10 inputs (resp. half the maximal result size in other experiments) for write rates $w$ and 1 input for cache miss differences for 11 inputs in total, followed by 2 hidden layers with each 30 neurons using rectified linear unit activations. Updates were performed using an Adam \citep{KingmaBa2014} updater with mini-batches of size 10 (all training was executed by the CPU due to the small model size). Learning was non-episodic and fully on-policy after an initial exploration period. The learning rate was set to $\alpha=0.0005$ with gradients clipped at 30, allowing for aggressive updates. 
\vspace{-2mm}
\subsection{Results}
\begin{figure}  [ht!]
    \centering
    \begin{subfigure}[t]{0.31\columnwidth}
        \centering
        \includegraphics[width=\columnwidth, keepaspectratio]
        {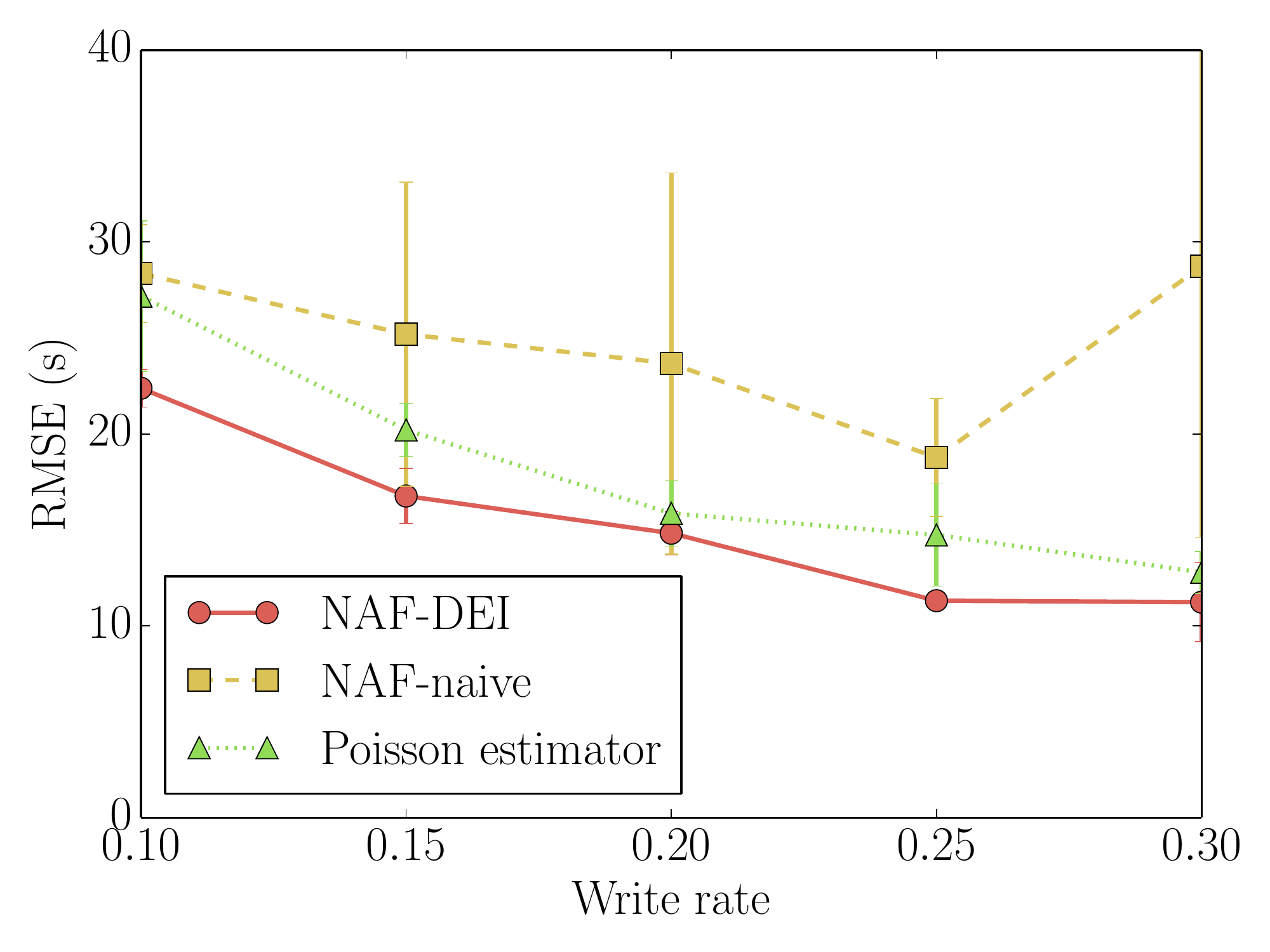}
        \caption{Errors against optimal policy for different workload mixtures.}
        \label{fig:write_rate_comparison}      
    \end{subfigure}
    \quad
        \begin{subfigure}[t]{0.31\columnwidth}
        \centering
        \includegraphics[width=\columnwidth, keepaspectratio]
        {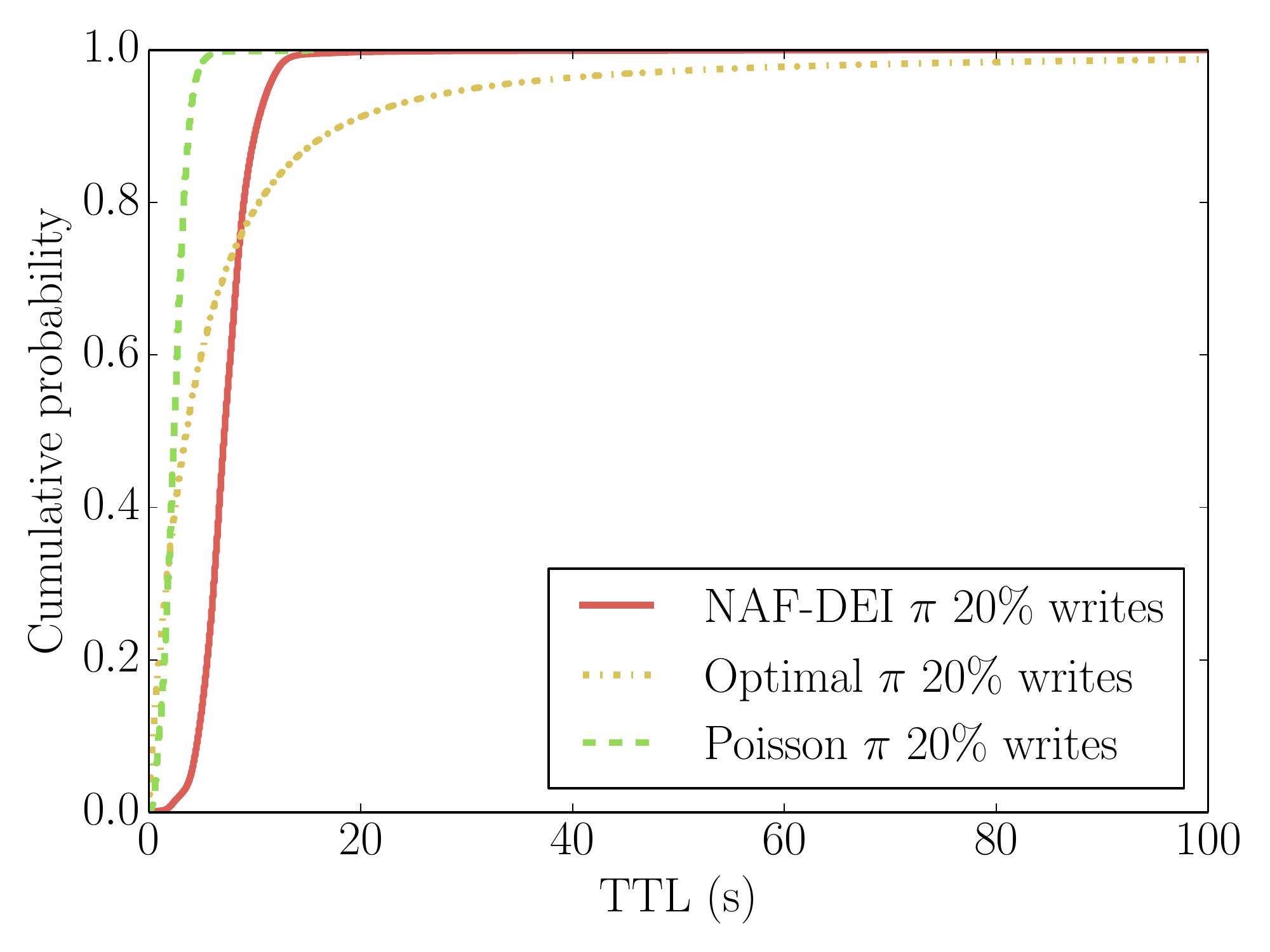}
        \caption{Comparison of learned, optimal and Poisson CDF for $w=20\%$.}
        \label{fig:cdf_comparison}
    \end{subfigure}
    \quad
    \begin{subfigure}[t]{0.31\columnwidth}
        \centering
        \includegraphics[width=\columnwidth, keepaspectratio]
        {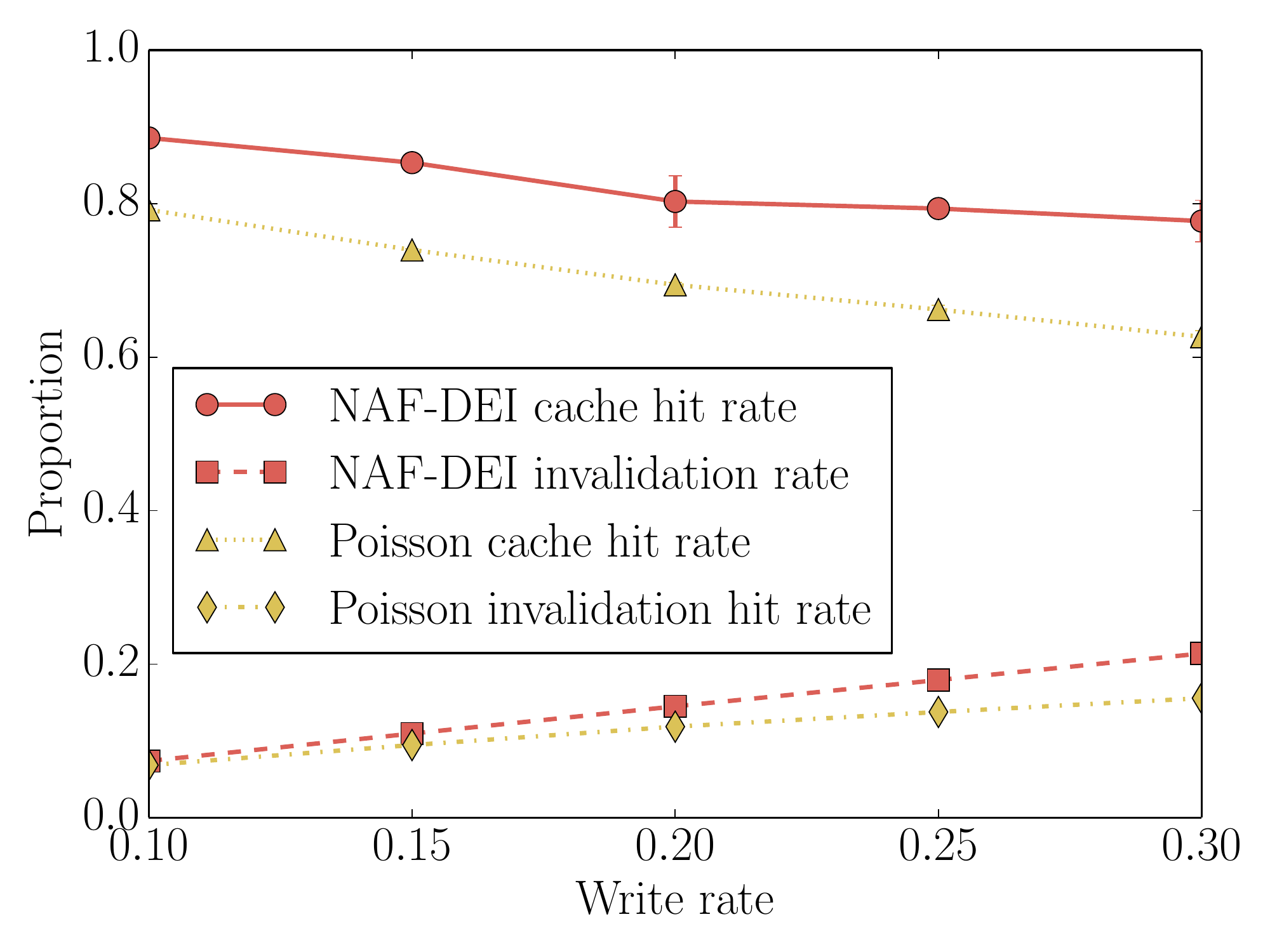}
        \caption{Cache performance.}
        \label{fig:cache_hit_comparison}      
    \end{subfigure}
    \caption{Evaluation of our approach against different baseline configurations.}
    \label{fig:writerate}
      \vspace{-2.5mm}
\end{figure}    
Figure \ref{fig:write_rate_comparison} compares results for different workload mixtures as the root-mean-squared per-step error (RMSE) against the optimal policy (truncated at the 99th percentile to remove outliers). The Poisson estimator is limited by its need to specify some default action if no write rate estimates are available. A feasible strategy is to set a maximum for TTLs and presume that the write rate on unknown objects corresponds to the inverse maximum. This is preferable to setting a single default estimate because such a default value would not account for different result sizes. We ran a number of configurations for the Poisson estimator and report the error for the best configuration (max TTL $300~s$). The NAF agent (using DEI) outperformed the Poisson estimator, as it is not dependent on a maximum value and uses 0 as input if no write rate is available instead of default per-key estimates. 

Further, we show the impact of running NAF without DEI (NAF-naive). NAF-naive instantly computes rewards and will thus rely on invalidations caused a by prior action, creating much larger error and larger standard deviations (for some configurations, learning from the wrong rewards leads to good accidental performance). NAF-DEI outperformed naive NAF by $38.5\%$ on average. While the TTL estimation problem is a special case due to the action being a time period where the delay is set to the action value, delayed asynchronous reward computation is likely beneficial to various other problems with high degrees of concurrency. 
We also observe that the absolute error is large for all solutions and improves with larger write rates, as more per-key information becomes available and the CDF turns steeper. Figure \ref{fig:cdf_comparison} compares CDFs from traces of NAF-learned, optimal and Poisson-estimated policies. Both approaches produce an overly steep CDF as the long tail of the optimal policy is principally difficult to predict.

It is important to recognize hat the theoretical optimal TTL we use to compute errors is not an ideal performance measure. It assigns an error to queries which expire without invalidation by computing the error until a future point in time when this query would have been invalidated if it had still been cached. In contrast, the reward function specifies a complex trade-off between invalidations, cache misses and global load. We nonetheless report this error since it allows for a more neutral comparison of strategies with different objective functions.
We also compared learning performance in different settings to a hypothetical default-value predictor which knows in advance which single default value would give the lowest error per workload mixture. Our results show that the learner can outperform the default-predictor by over $60\%$ for individual queries and up to $20\%$ for the top $20\%$ of queries (sorted by observed cache misses) for some workloads, performing better with smaller result sizes. With increasing write rates or increasing result sizes, true TTL distributions become more narrow and outperforming a default value is more difficult. However, the learner's mean error roughly matches (sometimes outperforms) the default-predictor due to large errors from the long tail of the access distribution. Note that the NAF-agent's error includes the online training period, as we wanted to evaluate the feasibility of a controller without prior training. 

Figure \ref{fig:cache_hit_comparison} shows actual cache performance in terms of achieved cache hit rates (higher is better) and invalidation rates (lower is better). NAF-DEI accepts slightly higher invalidation rates to ensure high cache performance while the Poisson estimator tends to predict lower TTLs due to using default write rates. Assuming an equal weighting between cache hit rate and invalidation rate, NAF-DEI offers much better cache performance with mean cache hit rate $88.5\%$ for $w=10\%$ with $7.3\%$ invalidations versus $79.5\%$ cache hit rate and $6.8\%$ invalidations for the Poisson estimator. For $w=30\%$, NAF-DEI achieved $77.7\%$ cache hits and $21.4\%$ invalidations versus $62.6\%$ cache hits and $15.6\%$ invalidations for the Poisson estimator. These results stress that simply estimating lower TTLs does not save many invalidations if estimates are imprecise on a per-query level.
\begin{figure}[ht!]
    \centering
    \begin{subfigure}[t]{0.31\columnwidth}
        \centering
        \includegraphics[width=\columnwidth, keepaspectratio]
        {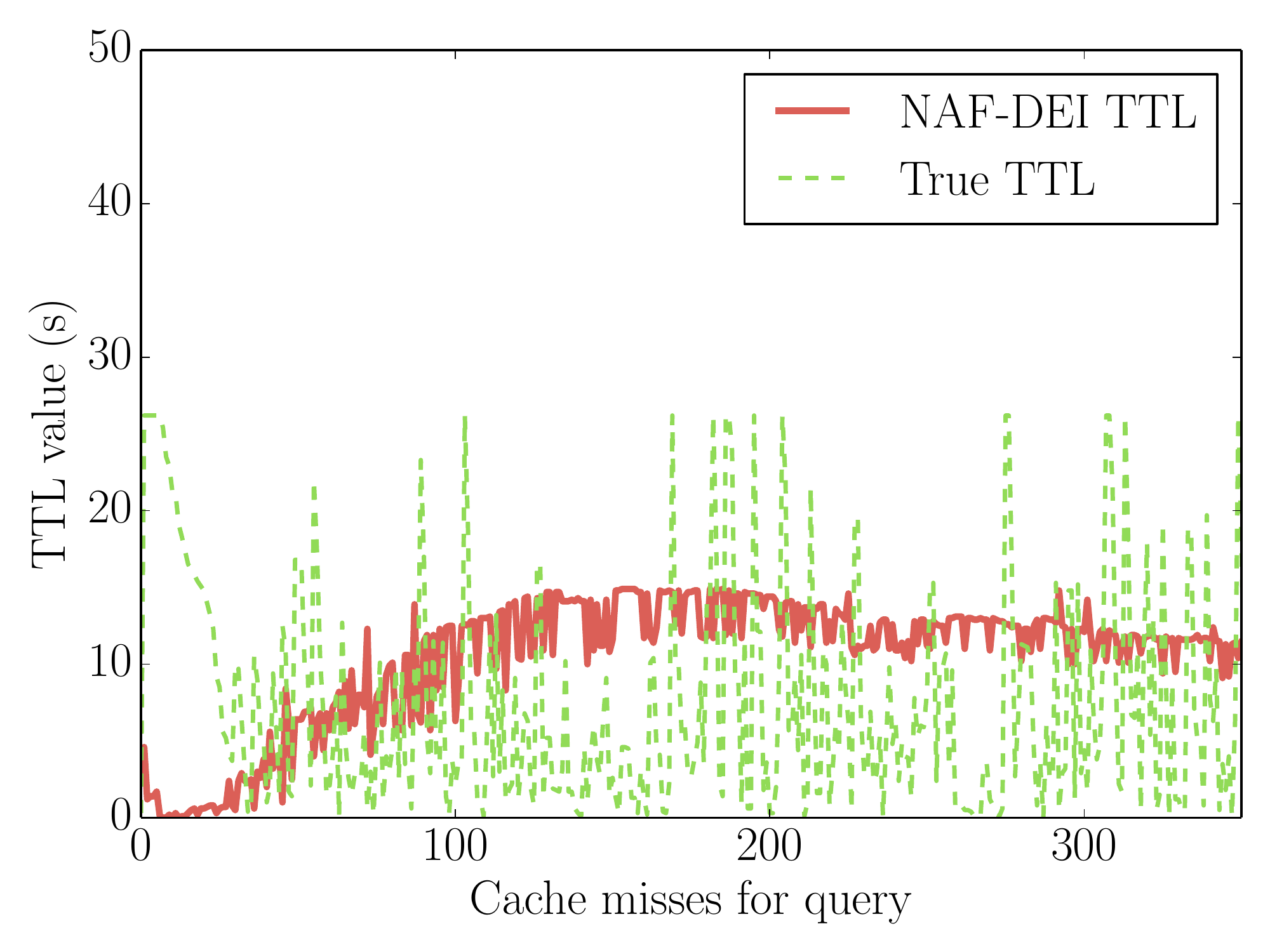}
    \caption{Query trace with $w=10\%$.}
        \label{fig:query_trace_10}      
    \end{subfigure}
    \quad
    \begin{subfigure}[t]{0.31\columnwidth}
        \centering
        \includegraphics[width=\columnwidth, keepaspectratio]
        {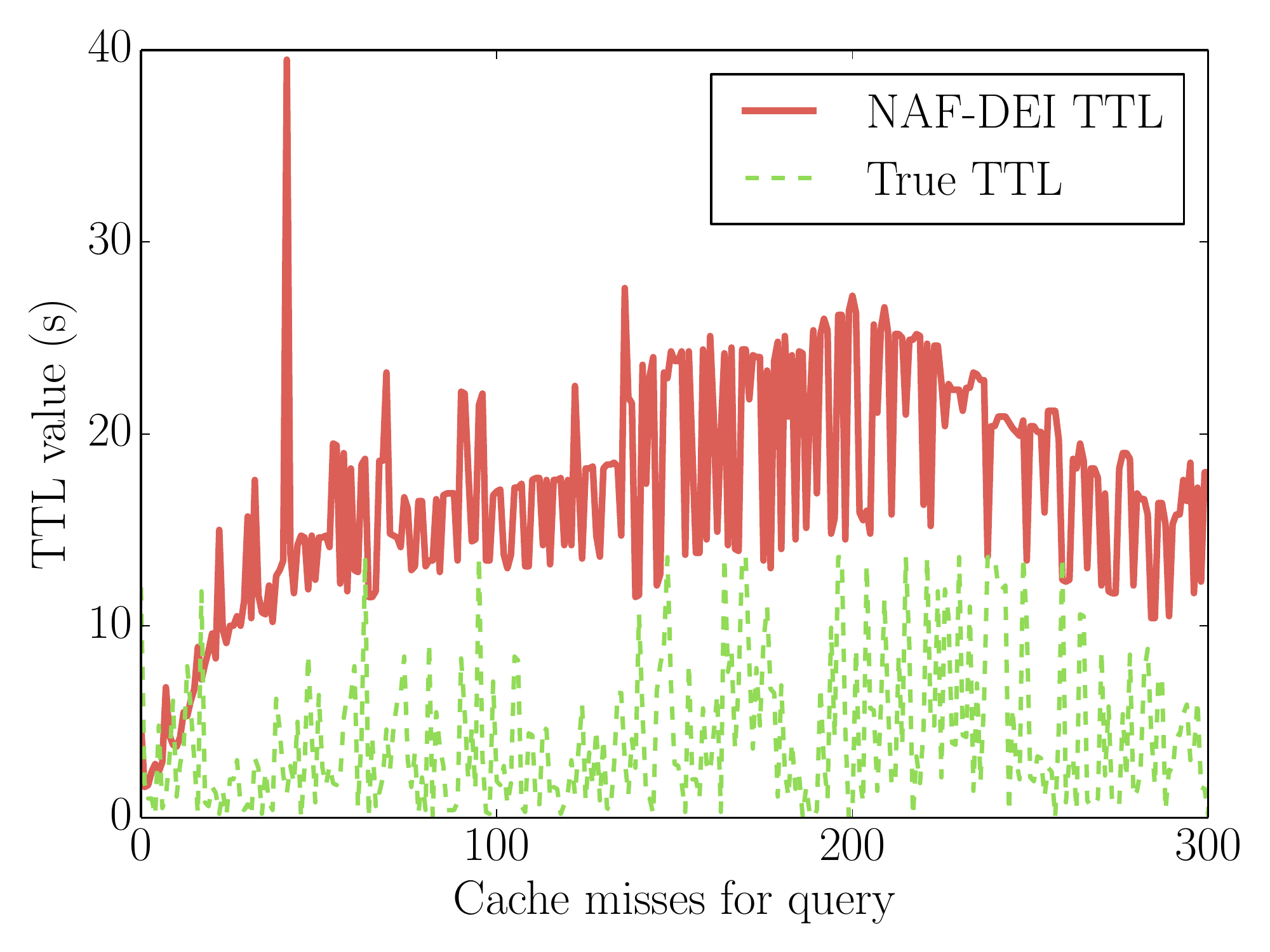}
        \caption{Query trace with $w=30\%$.}\label{fig:query_trace_30}
    \end{subfigure}
        \begin{subfigure}[t]{0.31\columnwidth}
        \centering
        \includegraphics[width=\columnwidth, keepaspectratio]
        {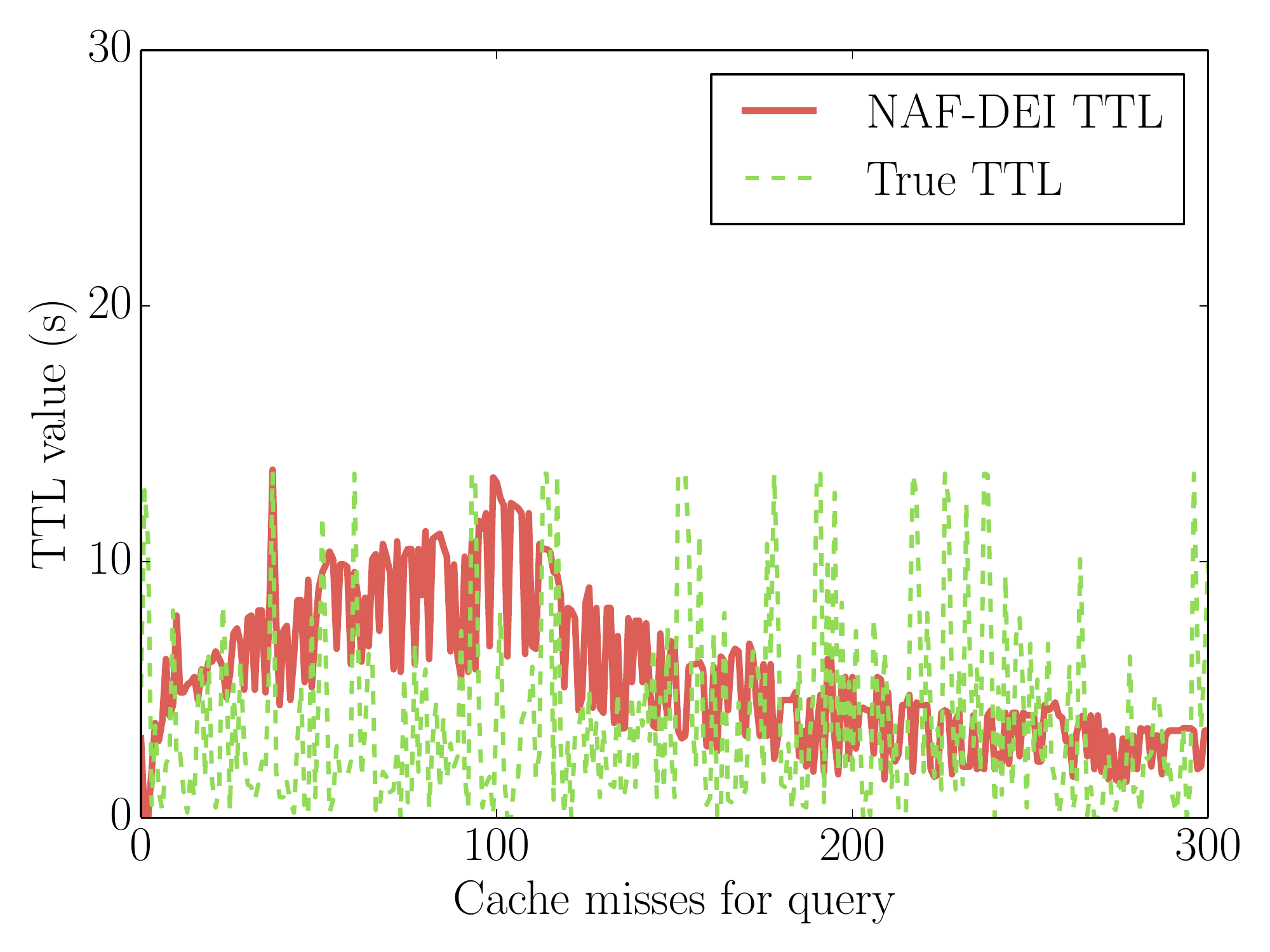}
        \caption{Reward-adjusted query trace with $w=30\%$.}\label{fig:query_trace_30_adjusted}
    \end{subfigure}
    \caption{Individual query trace examples illustrate learning over time and noise in the optimal policy.}
    \label{fig:query_traces}
   \vspace{-4mm}
\end{figure}

To better understand learning behavior, we examine traces of individual queries through experiment runs. Figure \ref{fig:query_trace_10} shows learned and optimal actions for every instance an individual query is observed throughout one experiment with $w=10\%$, illustrating both the online learning process and the noise in the optimal policy. Note that while learning seems fast, the plot does not show how much time (and hence learning from other queries) passes between each instance of the query. For a throughput of $1,000$ concurrent requests per second, most learning (by magnitude of error) took place in the first few minutes (about $20\%$ of experiment duration). For $w=30\%$, the agent uses the additional information from both more writes and faster changing cache miss rates to make more specific guesses, as seen in figure \ref{fig:query_trace_30}. While the learner seems to match the general shape of the series of true TTLs, it systematically overestimates true values by $5-10$ s. This is because the reward was statically configured to always encourage high cache hit rates independent of higher write rates. In figure \ref{fig:query_trace_30_adjusted}, we adjusted rewards according to the workload mixture, i.e. we encouraged the load to stay below $1 - w$. Consequently, TTLs for the query decrease over time once the cache fills up. 
The Poisson estimator could be similarly tuned by using not the expected value until seeing the next write but some other quantile, e.g. $75\%$ to encourage higher cache hit rates. However, RL-based solutions offer a natural interface to incorporate additional performance metrics without having to translate them into an analytical model, i.e. by determining which Poisson parameters correspond to the desired performance.

Our results allow some outlook on learning parameters at a much larger scale. Due to the Zipf nature of web workloads \citep{BreslauCaoFanEtAl1999}, most queries and updates will concentrate on a small sets of "hot" database records for which it might be feasible to track runtime information and use them for specific predictions. In conclusion, the combination of small model size and a high degree of concurrency allowed NAF-DEI to achieve an effective trade-off between avoiding invalidations and ensuring high cache hit rates without requiring prior training.
\section{Conclusion and future work}
To the best of your knowledge, this work presents the first application of deep reinforcement learning in predicting request-level parameters in computer systems. We introduced the concept of delayed experience injection to capture asynchronous reward/next-state semantics in concurrent environments where relevant metrics are only available later. The key idea of our work is that instead of learning global parameters from global metrics, DRL can facilitate per-request decisions based on fine-grained metrics. Results show that our NAF-based approach can outperform a statistical estimator on the TTL estimation problem by leveraging available runtime information. 

In future work, we will explore in more detail how to relate dynamic reward adjustments to specific service level objectives in non-stationary environments. We have also made the simplifying assumption of a single node backend receiving all incoming requests. Future work in this domain hence needs to investigate coordination and distributed learning in infrastructures where each node only observes part of the environment, as opposed to each node observing a separate copy of the problem.

\section*{Acknowledgements}
This work was supported by the EPSRC (grant reference EP/M508007/1) and a Computer Laboratory Premium Scholarship (Sansom scholarship).

\end{document}